%% file: main.tex
\documentclass[11pt]{article}
\pdfoutput=1
\usepackage{amssymb}

\usepackage[final]{acl}
\usepackage{times}
\usepackage{latexsym}
\usepackage[T1]{fontenc}
\usepackage[utf8]{inputenc}
\usepackage{wrapfig} 
\usepackage{graphicx}
\usepackage{amsmath}
\usepackage{amssymb}
\usepackage{mathtools}
\usepackage{amsthm}
\usepackage{tabularx}
\usepackage{float}
\usepackage{microtype}
\usepackage{subfigure}
\usepackage{bm}
\usepackage{bbm}
\usepackage{textcomp}
\usepackage{multirow}
\usepackage{booktabs}
\newcommand{\cellformat}[1]{\begin{tabular}{@{}p{\linewidth}@{}}#1\end{tabular}}
\usepackage{authblk}
\usepackage{xspace}
\usepackage{colortbl}
\usepackage{comment}
\usepackage{thmtools, thm-restate}
\usepackage{verbatim}
\usepackage{makecell}
\usepackage{array}
\usepackage{pifont}
\usepackage[compact]{titlesec}
\usepackage[subtle, mathdisplays=tight, charwidths=tight, leading=normal]{savetrees}
\usepackage{tikz}
\usetikzlibrary{arrows}
\usetikzlibrary{positioning}
\definecolor{lightblue}{HTML}{D9EAF7} 
\definecolor{williamblue}{RGB}{0, 102, 204} 

\usepackage[most]{tcolorbox}

\tikzset{
  treenode/.style = {align=center, inner sep=0pt, text centered,
    font=\sffamily},
  arn_n/.style = {treenode, circle, black, font=\sffamily\bfseries, draw=black,
    fill=white, text width=1.5em},
  arn_r/.style = {treenode, circle, black, font=\sffamily\bfseries, draw=black,
    fill=white, text width=1.0em},
  arn_x/.style = {treenode, rectangle, draw=black,
    minimum width=0.5em, minimum height=0.5em}
}

\usepackage{hyperref}

\usepackage{enumitem}

\newcommand*{\ShowNotes}{}
\input{macros.tex}
\newif\ifarxiv
\title{\lmunit: Fine-grained Evaluation with Natural Language Unit Tests}

\author{
\textbf{Jon Saad-Falcon}\textsuperscript{1,2}\thanks{ Co-first author.}, \textbf{Rajan Vivek}\textsuperscript{1*}, \textbf{William Berrios}\textsuperscript{1*} \\
\textbf{Nandita Shankar Naik}\textsuperscript{1}, \textbf{Matija Franklin}\textsuperscript{1}, 
\textbf{Bertie Vidgen}\textsuperscript{1} \\
\textbf{Amanpreet Singh}\textsuperscript{1}, \textbf{Douwe Kiela}\textsuperscript{1,2}, \textbf{Shikib Mehri}\textsuperscript{1} \vspace{0.4cm} \\
\textsuperscript{1}Contextual AI\\
\textsuperscript{2}Stanford University, Department of Computer Science \vspace{0.4cm}\\
\texttt{jonsaadfalcon@stanford.edu},
\texttt{shikib@contextual.ai}
}
\newcommand{\lmunit}{\textsc{LMUnit}}
\usepackage[ruled,vlined,linesnumbered]{algorithm2e}

\begin{document}

\maketitle

\input{sections/abstract}
\input{sections/introduction}

\input{sections/related_work}
\input{sections/methodology}
\input{sections/experiments}

\input{sections/discussion}

\input{sections/conclusion}
\input{sections/limitations}

\bibliography{main}

\appendix

\input{sections/appendix}

\end{document}

%% file: macros.tex
\ifdefined\ShowNotes
  \newcommand{\colornote}[3]{{\color{#1}\bf{#2 #3}\normalfont}}
\else
  \newcommand{\colornote}[3]{}
\fi

\definecolor{darkred}{rgb}{0.7,0.1,0.1}
\definecolor{darkgreen}{rgb}{0.1,0.5,0.1}
\definecolor{cyan}{rgb}{0.7,0.0,0.7}
\definecolor{dblue}{rgb}{0.2,0.2,0.8}
\definecolor{maroon}{rgb}{0.76,.13,.28}
\definecolor{burntorange}{rgb}{0.81,.33,0}
\definecolor{royalpurple}{rgb}{0.47,.31,0.66}

\ifdefined\ShowNotes
  
\else
  
\fi

%% file: sections/abstract.tex
\begin{abstract}
As language models become integral to critical workflows, assessing their behavior remains a fundamental challenge -- human evaluation is costly and noisy, while automated metrics provide only coarse, difficult-to-interpret signals. We introduce \textbf{natural language unit tests}, a paradigm that decomposes response quality into explicit, testable criteria, along with a unified scoring model, \textbf{\lmunit{}}, which combines multi-objective training across preferences, direct ratings, and natural language rationales.
Through controlled human studies, we show this paradigm significantly improves inter-annotator agreement and enables more effective LLM development workflows.
\lmunit{} achieves stateof-the-art performance on evaluation benchmarks including FLASK, BigGenBench, and RewardBench 2, while maintaining competitive results on the original RewardBench. These results validate both our proposed paradigm and scoring model, suggesting a promising path forward for language model evaluation and development. Our code has been released at
\url{https://github.com/ContextualAI/LMUnit}
with an MIT license.
\end{abstract}

%% file: sections/introduction.tex
\section{Introduction}
\label{sec:introduction}
The evaluation of generative language models remains one of the most fundamental challenges in natural language processing \cite{jones1995evaluating,deriu2021survey,smith2022human,chang2024survey} -- it determines how we measure progress and shapes the field's trajectory. As these models transition from research prototypes to production systems, users increasingly rely on them for critical workflows \citep{lin2024wildbench}, creating an urgent need for evaluation methods that identify response strengths/weaknesses, ensure reliability, and prevent costly regressions.
Yet current approaches fall short: human evaluation is expensive and struggles to discern subtle differences among top models \citep{hosking2023human, clark2021all, karpinska2021perils}, while automated metrics compress response quality into coarse scores \citep{stent2005evaluating,liu2016not} that rely on implicitly learned, often biased criteria \citep{dubois2024length, shankar2024validates, zhang2024comprehensive}. As models become more deeply integrated into essential workflows, it is imperative that our evaluation methodologies evolve in tandem, empowering LLM practitioners to reliably \textbf{detect subtle failures}, meaningfully \textbf{distinguish among top-performing systems}, and \textbf{generate actionable insights} that drive improvements.

\begin{figure}[t]
    \centering
    \includegraphics[width=1.0\linewidth]{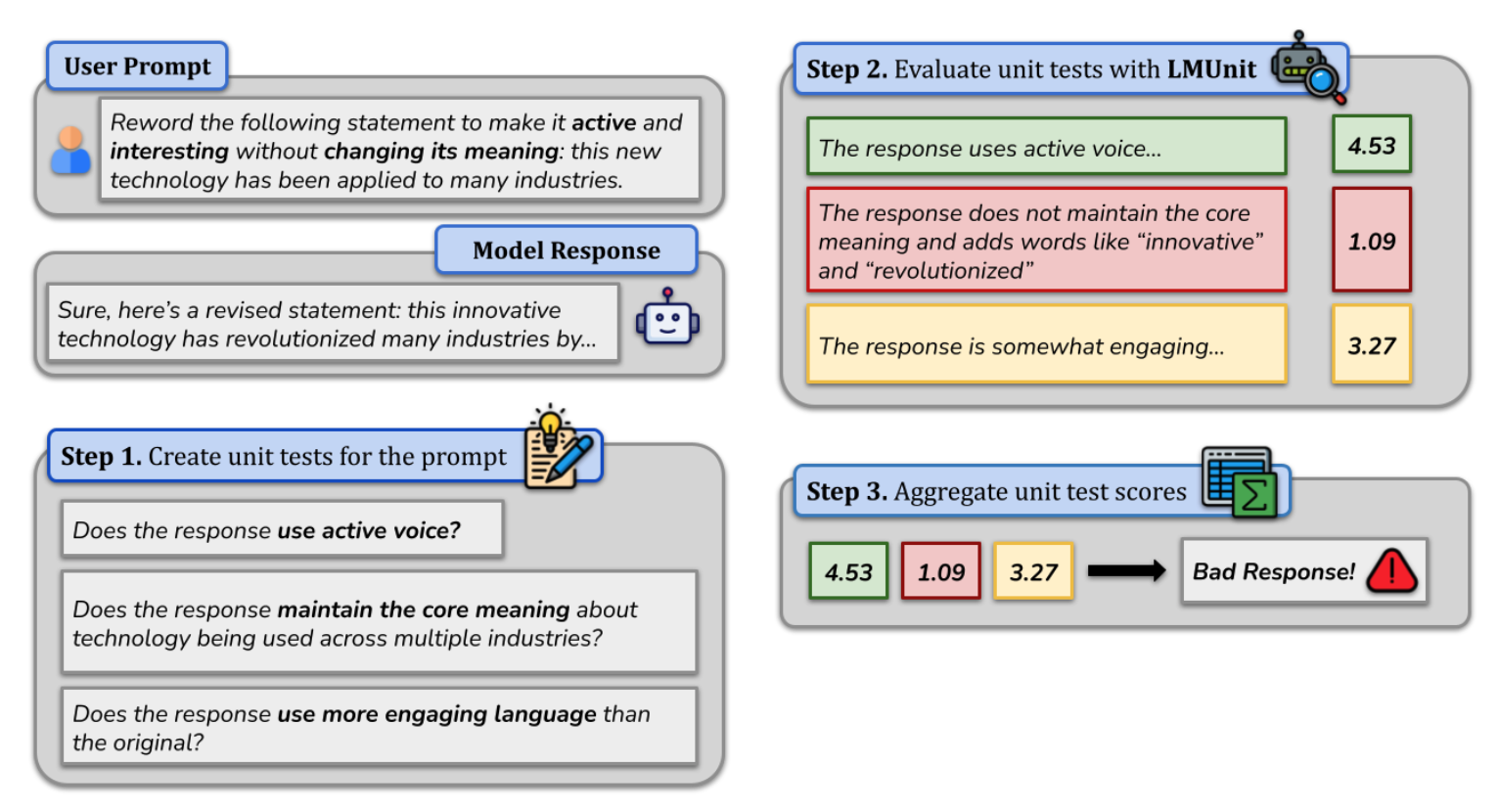}

    \caption{\textbf{Natural Language Unit Tests:} Overview of the three-step process: (1) unit test creation, (2) LMUnit-based scoring with natural language rationales, and (3) score aggregation for overall quality assessment.}
\vspace{-1.em}
    \label{fig:lmunit_overview}
\end{figure}

We focus on measuring response quality - one of the most critical challenges in evaluating language models. Defining ``response quality'' is inherently complex, depending on multiple factors including factual accuracy, logical coherence, and alignment with user objectives, which vary by domain, application, style, and context \citep{mehri2020unsupervised, ye2023flask, krishna2023longeval}. Existing approaches struggle with this complexity: (1) reference-based comparisons fail in open-ended scenarios where no single ``correct'' response exists \citep{liu2016not, lowe2017towards}, (2) human evaluations become inconsistent as models grow more capable and errors subtler \citep{walker2007individual, pan2024human, christiano2023deepreinforcementlearninghuman}, and (3) preference models and LLM judges compress nuanced assessments into opaque metrics that are difficult to interpret or steer \citep{dubois2024alpacafarm, d2024anchored, singhal2023long}. To address these limitations, we propose \textbf{natural language unit tests}, a paradigm that decomposes response quality into explicit, testable criteria that humans can define, refine, and guide over time (Figure \ref{fig:lmunit_overview}). While this approach enhances transparency, reliably scoring and integrating these fine-grained assessments while maintaining human values alignment remains a key challenge.

Building an effective scoring model for unit tests presents a significant challenge: it must accurately evaluate a wide range of criteria -- ranging from broad notions of quality to detailed rubrics that capture intricate, context-specific requirements. Existing approaches each address part of the problem: prompted LLM judges can be instructed to consider certain criteria \citep{liu2023g}, but their accuracy is limited by generic instruction-following abilities and the inability to learn directly from preference data \citep{wang2024direct, zhong2022towards}; preference models, while closely aligned with human judgments, lack promptability and struggle to handle more granular, human-defined criteria \citep{singhal2023long, lambert2023alignment}. 

To address these challenges, we propose \lmunit{}, a unified modeling approach that optimizes large language models as preference models while supporting flexible, user-defined evaluation criteria. By combining diverse training signals with natural language rationales, \lmunit{} achieves strong results across preference modeling, direct scoring, and fine-grained unit test evaluations, laying a robust foundation for more adaptive and transparent evaluation methodologies. These rationales are optional at inference time but enabling them allows further interpretability of scores.

To demonstrate our paradigm's effectiveness in enabling human stakeholder intervention, we assess its real-world impact through human studies. In a controlled annotation study, expert raters achieved higher inter-annotator agreement when evaluating outputs against explicit unit tests compared to standard preference annotations. Additionally, in a case study with LLM developers, \lmunit{}'s transparent, test-driven evaluations enabled identification of more errors than conventional LLM judges, demonstrating the value of our proposed paradigm

Our key contributions include: (1) proposing the paradigm of natural language unit tests, and validating it at scale, (2) developing \lmunit{} as a unified scoring model that achieves state-of-the-art performance, (3) showing the benefits and challenges of effective unit test creation and weighting strategies, (4) demostrating the importance of rationales when incorporating them as part of the training data. (5) validating our approach through human studies that demonstrate improved inter-annotator agreement and more effective LLM development workflows.

%% file: sections/related_work.tex
\section{Related Work}
\label{sec:related_work}

\subsection{Evaluation of Generative Language Models}
While human evaluation remains the gold standard for LLMs \citep{ouyang2022training, touvron2023llama}, its scalability limitations \cite{hosking2023human, schoch2020problem} have driven the development of automated approaches. These include word overlap metrics \citep{papineni2002bleu, lin2004rouge}, embedding-based scoring \citep{yuan2021bartscore, zhang2019bertscore}, model-based evaluations \citep{lowe2017towards, mehri2020usr, zhong2022towards, saad2023ares}, reward modeling \citep{christiano2017deep,askell2021general,kim2023aligning}, and LM judges \citep{zheng2023evaluation, liu2023g, es2023ragas, ravi2024lynxopensourcehallucination, kim2024prometheusinducingfinegrainedevaluation, li2024leveraging}. However, automated methods often lack interpretability and can show biases that diverge from human judgments \citep{shankar2024validates, wang2023large, chaudhari2024rlhfdecipheredcriticalanalysis}. Recent work has focused on developing fine-grained evaluators \citep{ye2023flask, wang2024direct, ribeiro-etal-2020-beyond, lin2023llm, cook2024tickingboxesgeneratedchecklists} and unifying evaluation paradigms \citep{wang2024direct, kim2024prometheus2opensource, wu2023fine}. For code generation specifically, LLM-based unit test generation has improved performance evaluation through compiler-compatible synthetic tests \citep{chen2022codet, yuan2023no, saad2024archon}.

\subsection{LM Judges}

LLMs can be prompted to evaluate responses without additional training, showing high correlation with human ratings \citep{liu2023g, wang2023chatgpt, fu2023gptscore, chiang2023can, es2023ragas, kocmi2023large, li2024generationjudgmentopportunitieschallenges}. While some approaches focus on in-context examples and evaluation instructions \citep{fu2023gptscore}, others leverage chain-of-thought prompting \citep{liu2023g} or fine-tune specialized judges \citep{saad2023ares, tang2024minicheckefficientfactcheckingllms}. However, these approaches face key limitations: poor generalization across evaluation tasks \citep{es2023ragas, saad2023ares, ravi2024lynxopensourcehallucination} and systematic biases in position, verbosity, and self-preference \citep{chen2024humans, pan2024human, zheng2023evaluation, koo2023cognitive}.

\begin{figure*}
    \centering
    \includegraphics[width=0.95\linewidth]{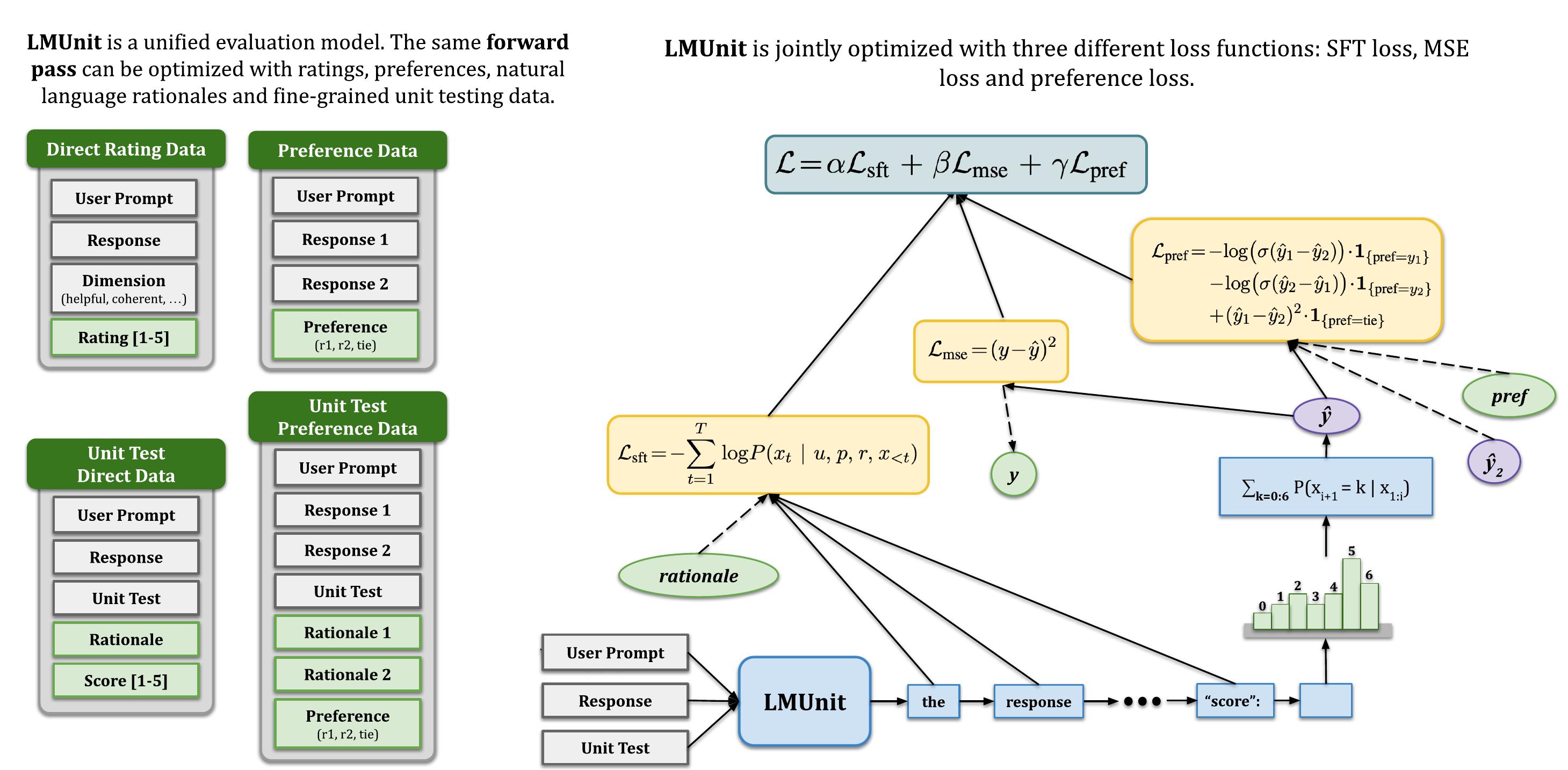}
    \caption{\textbf{\lmunit{} Training Setup}:
    We leverage several different data sources (direct rating, preference, unit test direct, unit test preference) along with three different loss functions, to optimize the fine-grained scoring of \lmunit{}.
    }
    \label{fig:lmunit_training_methodology}
\vspace{-1em}
\end{figure*}

\subsection{Reward Models}

Reward models, while widely adopted for evaluating and aligning language models \citep{bradley1952rank, christiano2017deep, skyworkreward2024}, face fundamental challenges: low inter-annotator agreement (65\% - 75\% - early RLHF papers) in human preference data \citep{askell2021general,ouyang2022training, wang2024secretsrlhflargelanguage}, noisy and inconsistent preferences \citep{dubois2024alpacafarm}, and spurious correlations like favoring longer responses \citep{lambert2023alignment, singhal2023long, dubois2024length}. Recent work shows promise in addressing these: Helpsteer-2 \citep{wang2023helpsteermultiattributehelpfulnessdataset} improved performance through better preference data collection. GenRM-COT \citep{zhang2024generativeverifiersrewardmodeling} and EvalPlanner \citep{saha2025learningplanreason} used chain-of-thought reasoning for more reliable evaluation. However, challenges with reward underspecification and alignment persist \citep{eisenstein2023reward, chaudhari2024rlhfdecipheredcriticalanalysis}.

\subsection{Fine-Grained Evaluators}

Breaking down complex evaluation problems has been foundational in NLP \citep{walker2000towards} and remains vital for language models \citep{saha2024branchsolvemergeimproveslargelanguage}. While early approaches used fixed evaluation dimensions \citep{liu2016not, lowe2017towards, zhong2022towards}, modern language models enable more dynamic, fine-grained criteria \citep{mehri2020unsupervised, lin2023llm, ye2023flask, kim2024biggenbenchprincipledbenchmark}, though pre-defined criteria may not generalize well to real-world settings \citep{shankar2024validates}. Our work builds upon CheckList \citep{ribeiro-etal-2020-beyond}, which introduced structured behavioral testing for NLP models, TICK \citep{cook2024tickingboxesgeneratedchecklists}, which demonstrated decomposition benefits through model-generated criteria, and CheckEval \citep{lee2025checkevalreliablellmasajudgeframework}, which showed that using a decomposition list of binary questions can effectively improve the average agreement across evaluator models and also reduce the score variance for text-generation tasks. We extend these approaches by training a dedicated scoring model that synthesizes multiple training signals, conducting broader evaluations across diverse benchmarks, and validating through human studies.

We have further discussion of how \lmunit{} relates to recent work in Appendix \ref{sec:more_related_work}
\subsection{Unified Evaluators}

Recent work has focused on unifying different evaluation paradigms. DJPO \citep{wang2024direct} improves human correlation by training LM judges through preference optimization \citep{rafailov2023direct}, while Prometheus \citep{kim2024prometheusinducingfinegrainedevaluation, kim2024prometheus2opensource} combines direct assessment and pairwise ranking capabilities through model weight merging. These approaches, along with fine-grained reward functions \citep{wu2023fine}, show promise in both human and automatic evaluations.

\lmunit{} extends these unified approaches while addressing limitations in interpretability, generalization, and fine-grained control. It decomposes evaluation into explicit testable criteria defined and refined by human experts, leveraging both LM judges (natural language understanding, flexible criteria) and reward models (precise scoring, preference learning) to enable reliable, interpretable, and actionable evaluation adaptable to diverse real-world requirements.

%% file: sections/methodology.tex
\section{\lmunit{} Methodology}
\label{sec:methodology}

To enable reliable scoring of natural language unit tests, we develop \lmunit{}, a unified modeling approach that combines multi-objective training with natural language rationale generation. The key challenge lies in effectively integrating diverse training signals while maintaining both high accuracy and interpretable outputs. Here, we detail our approach to addressing this challenge through careful problem formulation, synthetic data generation, and our training methodology.

\subsection{Problem Formulation}

The core challenge in language model evaluation is developing scoring models that can reliably evaluate responses against specific criteria while providing interpretable reasoning. Our formulation centers on unit tests: given a unit test $u$, prompt $p$, and response $r$, we train models to generate both rationales and scores through the mapping $f(u,~p,~r) \rightarrow {\text{rationale}, ~\text{score}}$.

Our approach builds on two existing forms of evaluation data:
direct rating data $(p,r) \rightarrow \text{score}$ and preference data $(p,r_1,r_2) \rightarrow \text{preference}$. We extend these into unit test-based formats:

\begin{enumerate}
\item Unit test direct data: $(u,~p,~r) \rightarrow \text{score}$ or $(u,~p,~r) \rightarrow ~{\text{rationale}, ~\text{score}}$
\item Unit test preference data: $(u,~p,~r_1,~r_2) \rightarrow \text{pref}$ or $(u,~p,~r_1,~r_2) \rightarrow ~{\text{rationale}_1, ~\text{rationale}_2, ~\text{pref}}$
\end{enumerate}

This formulation leverages two complementary data sources: naturally occurring preference and rating data to capture human preferences and calibrate against absolute quality scales, alongside synthetic data that enables fine-grained evaluation of specific criteria with interpretable rationales. At inference time, \lmunit{} can flexibly operate with or without rationale generation.

\subsection{Synthetic Data Pipeline}

Our data generation pipeline operationalizes the unit test formulation through three key stages, producing examples scored on a 1-5 scale where higher scores indicate better satisfaction of the criteria:
\begin{enumerate}
\item \textbf{Unit Test Generation}: For each prompt, we generate diverse unit tests targeting fine-grained quality criteria. To encourage focus on response-specific details, we optionally provide one or two responses during generation. We also maintain a set of coarse-grained global tests (see Table~\ref{tab:global-unit-tests} for details) to ensure broad coverage of general quality dimensions.
\item \textbf{Contrastive Response Generation}: For each $(u,~p,~r)$ triplet, we generate contrastive responses that vary systematically in how well they satisfy the unit test criteria. This creates rich training signal for learning fine-grained quality distinctions.
\item \textbf{Rationale and Score Generation}: For a subset of examples, we generate chain-of-thought rationales that explicitly reason through the evaluation criteria. Each rationale concludes with a score that must align with any existing seed data scores to maintain consistency.
\end{enumerate}
We seed our synthetic data pipeline with prompts, responses, tests and scores from diverse sources including Nectar \citep{zhu2024starling}, Prometheus \citep{kim2024prometheusinducingfinegrainedevaluation}, Tulu3 \citep{lambert2024t}, Complex Instructions \citep{he2024complex}, Infinity-Instruct \citep{InfinityInstruct2024}, and HelpSteer2 \citep{wang2024helpsteer2,wang2024helpsteer2preferencecomplementingratingspreferences}. 

\subsection{Training}
\label{sec:training}
LMUnit combines the strengths of generative judge models and classifier-based reward models through a unique multi-objective training approach. Given a unit test $u$, prompt $p$, and response $r$, the model outputs a sequence of rationale tokens $\mathrm{rat} = (\mathrm{rat}_1, ..., \mathrm{r}_T)$ followed by a score token $s$. The probability distribution over possible score values $k \in {0,1,\ldots,6}~$ is:
\begin{equation}
P(s = k ~ | ~ u, ~ p, ~ r, ~ \mathrm{rat}) = \text{softmax}(\mathbf{h}^T \mathbf{W}_s)_k
\end{equation}
We compute a continuous score prediction through a weighted sum:
\begin{equation}
\hat{y} = \sum_{k=0}^6 ~ k ~ \cdot P(s = k ~ | ~u, ~p, ~r, ~\mathrm{rat})
\end{equation} 

The training objective combines three losses. 
First, SFT loss on the rationale and score tokens: 
\begin{equation}
\mathcal{L}_{\text{sft}} = -\sum_{t=1}^T ~ \log P(x_t ~ \mid ~u, ~p, ~r, ~x_{<t})
\end{equation}
where $x_{1:t}$ represents tokens in both rationale and score sequences.

Second, MSE loss on the continuous score prediction:
\begin{equation}
\mathcal{L}_{\text{mse}} = (y - \hat{y})^2
\end{equation}

Third, preference loss:
\begin{align}
\mathcal{L}_{\text{pref}} &= -\log\bigl(\sigma(\hat{y}_1 - \hat{y}_2)\bigr) \cdot \mathbf{1}_{\{\text{pref}=y_1\}} \nonumber\\
&\quad -\log\bigl(\sigma(\hat{y}_2 - \hat{y}_1)\bigr) \cdot \mathbf{1}_{\{\text{pref}=y_2\}} \nonumber\\
&\quad +(\hat{y}_1 - \hat{y}_2)^2 \cdot \mathbf{1}_{\{\text{pref}=\text{tie}\}}
\end{align}

Here, $\sigma$ is the sigmoid function. The final loss is a weighted combination:
\begin{equation}
\mathcal{L} = \alpha\mathcal{L}_{\text{sft}} ~+ ~\beta\mathcal{L}_{\text{mse}} ~ +  ~\gamma\mathcal{L}_{\text{pref}}
\end{equation}

\subsection{Post-Training of Rationales}

While our initial model learns to generate rationales through imitation learning, there is no guarantee that these rationales actually improve scoring performance. We address this by collecting pairs of desirable and undesirable rationales for direct preference optimization \citep{rafailov2023direct}, training the model to prefer rationales that lead to correct scoring. We employ several collection strategies: Through Refined, we collect on-policy rationales from our trained model and use the teacher to refine them through revisions \citep{d2024anchored} that improve scoring accuracy. With Harmonized, we provide the teacher with two model rationales from a preference pair to harmonize them with their samples' relative quality. In the Teacher-based strategy, we sample teacher rationales on known-score samples, using those with correct outcomes as chosen samples and incorrect ones as rejected. We compare these approaches in Table \ref{tab:rationale_ablations}.

\subsection{Bayesian Optimization of Global Unit Tests}
\label{sec:bayesian-optimization}
Natural language unit tests decompose evaluation into fine-grained criteria through K global tests that assess dimensions like accuracy, safety, and coherence. The aggregation of these assessments into an overall score is crucial for valid evaluation. Rather than using standard uniform weighting, we learn optimal weights $w_1,...,w_K$ through Bayesian optimization over human preference data to maximize alignment between weighted test scores and human judgments. This process iteratively updates weights from a uniform initialization based on agreement with held-out human preferences.

%% file: sections/experiments.tex
\section{Experiments}
\label{sec:experiments}
\input{tables/lmunit_main_results}

We conducted extensive experiments to evaluate \lmunit{} and the natural language unit test paradigm. First, we evaluated the performance of \lmunit{} on several evaluation benchmarks, comparing to LLMs as judges, reward models, and trained evaluation models. Next, we perform ablations to understand the impact of different methodologies, including loss functions and data mixture choices. Also, we examined improving rationales through post-training and analyzed the impact of decomposition through several unit test strategies. Finally, as shown in Appendix \ref{sec:user_study}, we also conducted two human subject studies to validate the advantages of the \lmunit{} paradigms over LM judges

\subsection{Experimental Setup}
\subsubsection{Model Configuration and Training Data}
Our training data encompasses a diverse mix of preference judgments, direct scores, and rationales across multiple sources: (i) \textsc{Helpsteer 2} ($50$K pairs with ratings spanning five dimensions), (ii) \textsc{Prometheus} ($10$K unpaired samples with ratings), (iii) \textsc{Synth Non-Rubric} ($11$K pairs with ratings and rationales), (iv) \textsc{Synth Rubric} ($13$K unpaired samples with ratings and rationales).

We train several variants of \lmunit{} initialized from instruction-tuned LLaMa-3.1 models (8B, 70B). We train our models for 2000 steps using fixed weights (i.e., $\alpha = \beta = \gamma = 1$) for the different loss components, with a 5x loss multiplier applied to the rationale samples. The training uses the Adam optimizer \cite{kingma2015adam} with a learning rate of 1e-6 and a cosine learning rate scheduler, using a batch size of 64 and a sequence length of 8K. We estimate using 40k GPU hours on 8 8XH100 nodes for training experiments.

\subsubsection{Evaluation Benchmarks}
\label{sec:eval-benchamrks}
We evaluate our models on six benchmarks spanning diverse capabilities: Direct scores assessment (BigGenBench, Flask), Classification (Internal Unit Test set, Infobench), and preference evaluation (RewardBench , LFQA). We further evaluate on RewardBench 2 \cite{malik2025rewardbench2advancingreward} against an updated set of baselines in Table \ref{tab:model_comparison_rewardbench2}. At inference time, we compute a continuous score as the expected value of the possible scores in accordance to our training strategy described in Sec. \ref{sec:training}. For dataset details, see Appendix \ref{sec:eval-detail}

\subsection{Key Results}
Our models demonstrate strong performance across diverse evaluation settings (Table~\ref{table:model_comparison_results}). On direct assessment tasks, \lmunit{} achieves state-of-the-art results with correlations of \textbf{72.03} on FLASK and \textbf{67.69} on BiGGen-Bench, where fine-grained evaluation is particularly important. In aggregate, \lmunit{} achieves strong overall performance with scores of \textbf{79.74} (eight weighted global unit tests) and \textbf{79.29} (single unit test), outperforming general-purpose models like GPT-4 (78.29) and Claude-3.5 Sonnet (77.78). Even our smaller \lmunit$_{\text{LLaMA3.1-8B}}$ variant remains highly competitive with a 74.10 average score.
 For pairwise ranking tasks, using unweighted global unit tests slightly decreases overall performance to 78.78 (-$\mathbf{0.96}$), but \lmunit{} remains stronger than all other baselines. We recover this minor performance loss through Bayesian optimization of the global unit test weights while reaching \textbf{93.45} on RewardBench (+$\mathbf{2.91}$) - though we note this weighting is learned on a subset of RewardBench itself, analogous to tuning hyperparameters on the test set (following a similar experimental setup as \citet{wang2024helpsteer2}). A more rigorous analysis using a proper held-out evaluation set is provided in Section \ref{sec:decomp_test}, confirming the generalization of this method. Furthermore, on the more recent RewardBench 2 benchmark (designed to be substantially more difficult than the original), \lmunit{} achieves state-of-the-art performance and remains the best generative reward model as of September 2025. These strong results across direct assessment, classification, and pairwise ranking tasks validate the effectiveness of our synthetic data pipeline, training setup, and unified scoring methodology, establishing \lmunit{} as a state-of-the-art model for reliable evaluation.

\subsection{Ablation Studies}
We conduct extensive ablation studies to understand the key components driving \lmunit{}'s performance. Our analysis focuses on three main aspects: (1) the impact of different training objectives and data mixture compositions, (2) the role of rationales in model performance, and (3) strategies for unit test decomposition and aggregation. Additionally, we perform supplementary ablations on \lmunit{} such as base-model architecture (\ref{sec:base_model_arch}) unit test composition (\ref{sec:unit_test_composition}), Bayesian optimization with different models (\ref{sec:bayesian_optimization_ablation}), and \lmunit{} weighted inference (\ref{sec:custom_inference}).
\subsubsection{Impact of Loss Functions}
Our ablation studies in Table~\ref{tab:lmunit_evaluation_ablation} demonstrate that combining training objectives (SFT, MSE, and preference loss) yields measurable improvements across our evaluation benchmarks ($+0.5$). \lmunit{}-8B shows particularly significant gains on fine-grained evaluation datasets---9\% on FLASK and 6\% on BigGenbench---with more modest improvements (1-3\%) on pairwise datasets that assess coarser-grained capabilities. These differential gains suggest our multi-objective approach is especially beneficial when evaluating nuanced LLM capabilities and when parametric capacity is limited, as evidenced by smaller improvements ($+3\%$) at the 70B parameter scale.
\input{tables/methods_ablation}
\vspace{-0.5em}
\subsubsection{Data Mixture Effects}
We analyze how different compositions of training data affect \lmunit's performance to identify the most effective mixture for robust evaluation capabilities. As shown in Table~\ref{tab:data_mix}, rubric data is essential for strong performance on fine-grained direct assessment and that our synthetic data pipeline provides dramatic performance gains (+$\mathbf{3.52}$) when synthetic rubric data is incorporated. We also observe that non-rubric synthetic data is most effective as preference pairs (+$\mathbf{4.04}$) rather than direct scoring data (-$\mathbf{2.75}$), likely due to the improved contrastive signal.

\input{tables/data_mixes}
\subsubsection{Impact of Rationales}
Moving beyond simple imitation learning of rationales, we examine strategies to optimize rationale generation for better evaluation. As shown in Table~\ref{tab:rationale_ablations}, training with rationales improves model performance even when rationales are not used at test time (+$\mathbf{0.2}$). While including rationales during inference initially leads to lower scores, our post-training optimization through DPO helps recover performance, with teacher-based pairs providing the largest gains (+$\mathbf{1.1}$). 
\input{tables/rationales_ablations}

\subsubsection{Unit Test Decomposition Analysis}
\label{sec:decomp_test}
Our experiments with different unit test strategies on RewardBench (Table~\ref{tab:decomposition}) reveal two key findings. First, global-level tests significantly outperform query-level tests across all categories, with section-level learned weights achieving the strongest results (+$\mathbf{2.4}$ over unweighted aggregation). Second, the performance of fine-grained query-level tests degrades substantially, particularly on harder examples, though this can be partially mitigated by placing greater weight on earlier tests (+$\mathbf{1.5}$).

These results highlight both the promise and challenges of our approach: while global unit tests provide a robust foundation for evaluation, developing effective fine-grained testing criteria remains difficult. The success of weighted global unit tests, coupled with the challenges of query-level decomposition, suggests an important direction for future work in developing more sophisticated test generation and aggregation strategies. Additional details of how decomposition is applied with Bayesian optimization and with different base models can be seen at \ref{sec:bayesian_optimization_details}

%% file: tables/lmunit_main_results.tex
\begin{table*}[ht]
  \centering
  \small
  \resizebox{0.95\textwidth}{!}{%
    \setlength{\tabcolsep}{5pt}{
      \begin{tabular}{lccccccccc}
        \toprule
        \multirow{3}{*}{\bf Model} & 
        \multicolumn{2}{c}{\bf Direct Assessment} & 
        \multicolumn{2}{c}{\bf Classification} & 
        \multicolumn{2}{c}{\bf Pairwise Ranking} & 
        \multirow{3}{*}{\bf Average*} \\
        \cmidrule(lr){2-3}
        \cmidrule(lr){4-5} 
        \cmidrule(lr){6-7}
        & Flask & BiGGen-Bench & Human-Internal & 
        InfoBench & RewardBench & LFQA \\
        \midrule
        GPT-4o  & 69.00 & 65.00 & 81.80 & \textbf{92.80} & 84.60 & 76.54  & \cellcolor{lightblue}77.59 \\
        Claude-3.5 Sonnet  & 67.25 & 61.83 & 84.53 & 91.58 & 84.23 & \textbf{77.24} & \cellcolor{lightblue}76.43 \\
        \midrule                
        Prometheus-2-7B  & 47.00 & 50.00 & 75.58 & 48.60 & 72.0 & 72.31 & \cellcolor{lightblue}57.98 \\
        Prometheus-2-8x7B  & 54.00 & 52.00 & 77.82 & 87.85 & 74.5 & 74.23 & \cellcolor{lightblue}68.52 \\
        Prometheus-2-BGB-8x7B & 31.00 & 44.00 & 78.57 & 83.87 & 68.3 & 71.54 & \cellcolor{lightblue}59.74 \\
        Llama-3-OffsetBias-8B & 29.00 & 21.00 & 68.15 & 72.15 & 84.0 & 63.08 & \cellcolor{lightblue}53.85 \\
        Skywork-Critic-Llama-3.1-8B & - & - & - & - & 89.0 & 64.23 & \cellcolor{lightblue} - \\
        SFR-LLaMA-3.1-8B-Judge  & 52.00 & 59.00 & - & \textbf{92.80} & 88.7 & 68.85 & \cellcolor{lightblue}72.27 \\
        SFR-LLaMA-3.1-70B-Judge & 66.00 & 65.00 & - & 92.58 & 92.7 & 75.00 & \cellcolor{lightblue}78.26  \\        
        \midrule  
        \lmunit$_{\text{LLaMA3.1-8B}}$ & 60.02 & 64.46 & \textbf{94.14} & 91.26 & 83.23 & 71.54 & \cellcolor{lightblue}74.10  \\
        \lmunit$_{\text{LLaMA3.1-70B}}$ & \textbf{72.03} & \textbf{67.69} & 93.63 & 89.00 & 91.56 & 76.15  &  \cellcolor{lightblue}{79.29}  \\
        \lmunit$_{\text{LLaMA3.1-70B}-Decomposed}$ & \textbf{72.03} & \textbf{67.69} & 93.63 & 89.00 & 90.54 & 74.62  &  \cellcolor{lightblue}78.78  \\
        \lmunit$_{\text{LLaMA3.1-70B}-Decomposed-Weighted}$$~\dagger$ & \textbf{72.03} & \textbf{67.69} & 93.63 & 89.00 & \textbf{93.45} & 76.53  &  \cellcolor{lightblue}\textbf{79.74}  \\
        \bottomrule
      \end{tabular}
    }
  }
  \caption{\textbf{Comprehensive Model Performance Comparison}: Evaluation results across multiple benchmarks showing model performance on various tasks. Metrics: (i) Pearson correlation coefficient for direct assessment, (ii) binary accuracy for classification tasks, and (iii) pairwise preference accuracy for pairwise comparisons. $\dagger$ represents our result with Bayesian optimization over pairwise benchmarks for learning global unit test weights, as described in Section \ref{sec:bayesian-optimization}. We learned dataset-level weights for LFQA and section-level weights for RewardBench by optimizing over model predictions on a 50\% split of the dataset, following prior work \citep{wang2024helpsteer2}. We only apply the decomposed unit tests and weight optimization for RewardBench and LFQA since they lack fine-grained criteria for evaluation. We confirm that this technique generalizes to a held-out split of RewardBench in Table \ref{tab:decomposition}. Note that the Average column excludes Human-Internal scores in order to compare fairly against the non-public SFR-LLaMA baselines (as of December 2024). We further evaluate on RewardBench 2 against more recent baselines in \ref{tab:model_comparison_rewardbench2}}
  \label{table:model_comparison_results}
\vspace{-2.0em}
\end{table*}

%% file: tables/methods_ablation.tex
\begin{table*}[!ht]
\centering
\small
\setlength{\tabcolsep}{5pt}%
\resizebox{0.95\textwidth}{!}{%
\begin{tabular}{lcccccccccc}
\toprule
\multirow{3}{*}{\bf Training Loss} & 
\multicolumn{2}{c}{\bf Direct Assessment} &
\multicolumn{2}{c}{\bf Classification} &
\multicolumn{2}{c}{\bf PairWise Ranking} & \multirow{3}{*}{\bf Average} \\
\cmidrule(lr){2-3}\cmidrule(lr){4-5}\cmidrule(lr){6-7}
& Flask & BiGGen-Bench & Human-Internal & InfoBench & RewardBench & LFQA \\
\midrule
\textbf{\lmunit$_{\text{LLaMA3.1-8B}}$}\\
\midrule
\textsc{SFT} & 51.31 & 59.12 & 94.19 & 90.29 & 83.56 & 68.85 & \cellcolor{lightblue}74.55 \\
\textsc{SFT + MSE} & 60.46 & 63.94 & 94.29 & 92.92 & 83.44 & 71.54 & \cellcolor{lightblue}\textbf{77.77} \\
\textsc{SFT + MSE + PREF} & 60.02 & 64.46 & 94.14 & 91.26 & 83.23 & 71.54 & \cellcolor{lightblue}77.44 \\
\midrule
\textbf{\lmunit$_{\text{LLaMA3.1-70B}}$}\\
\midrule
\textsc{SFT} & 69.09 & 67.14 & 93.88 & 90.83 & 89.98 & 76.15 & \cellcolor{lightblue}81.18 \\
\textsc{SFT + MSE} & 70.25 & 67.34 & 93.73 & 87.59 & 91.03 & 75.77 & \cellcolor{lightblue}80.95 \\
\textsc{SFT + MSE + PREF} & 72.03 & 67.69 & 93.63 & 89.00 & 91.56 & 76.15 & \cellcolor{lightblue}\textbf{81.68} \\
\bottomrule
\end{tabular}%
}
\caption{\textbf{Training Loss Ablation Results}: Adding SFT, MSE, and preference loss components each contribute modest but consistent improvements to \lmunit{}'s performance across direct assessment (Pearson correlation), classification (binary accuracy), and pairwise ranking (preference accuracy) tasks.}
\vspace{-0.5em}
\label{tab:lmunit_evaluation_ablation}
\end{table*}

%% file: tables/data_mixes.tex
\begin{table*}
\centering
\resizebox{0.95\linewidth}{!}{%
\setlength{\tabcolsep}{5pt}{
\begin{tabular}{lcccccccccc}
\toprule
\multirow{3}{*}{\bf Data Mix} & 
\multicolumn{2}{c}{\bf Direct Assessment} &
\multicolumn{2}{c}{\bf Classification} &
\multicolumn{2}{c}{\bf PairWise Ranking} & \multirow{3}{*}{\bf Average} \\
\cmidrule(lr){2-3}
\cmidrule(lr){4-5}
\cmidrule(lr){6-7}
& Flask & BiGGen-Bench & Human-Internal & InfoBench & RewardBench & LFQA \\
\midrule
\textit{\textbf{Direct only}}\\
\midrule
\textsc{HS2} & 57.0 & 42.26 & 94.74 & 88.60 & 91.31 & 69.23 & \cellcolor{lightblue}73.86\\
\textsc{HS2 + Synth Non-Rubric} & 47.00 & 42.00 & 93.83 & 88.80 & 86.00 & 69.00 & \cellcolor{lightblue}71.11 \\
\textsc{HS2 + Prometheus} & 64.90 & 59.27 & 93.43 & 87.50 & 91.40 & 71.15 & \cellcolor{lightblue}77.94 \\
\textsc{HS2+ Prometheus + Synth Rubric} & 71.60 & 67.94 & 94.89 & 89.19 & 91.70 & 73.50 & \cellcolor{lightblue}\textbf{81.46} \\
\midrule
\textit{\textbf{Preference only}}\\
\midrule
\textsc{Synth Non-Rubric} & 65.94 & 62.80 & 92.37 & 91.69 & 80.73 & 66.92 & \cellcolor{lightblue}76.74\\
\textsc{HS2}& 59.26 & 44.00 & 94.19 & 87.49 & 90.54 & 69.62 & \cellcolor{lightblue}{74.18}\\
\textsc{HS2 + Synth Non-Rubric}& 64.89 & 62.13 & 93.88 & 87.70 & 91.49 & 69.23  &  \cellcolor{lightblue}\textbf{78.22} \\
\midrule
\textit{\textbf{Full Data Mix}}\\
\textsc{All}& 72.03 & 67.69 & 93.63 & 89.00 & 91.56 & 76.15  &  \cellcolor{lightblue}\textbf{81.68} \\
\bottomrule
\end{tabular}
}}
\caption{\textbf{Training Data Mix Ablations}: Our direct-only synthetic mix with rubrics dramatically improves model performance over baselines trained on open-source data only. Our synthetic preference data also strongly improves performance even without rubrics, likely due to fine-grained contrastive signal. Training on our full data mix yields our SOTA \lmunit{} model. All models are initialized with Llama-3.1-70B. \textsc{HS2} refers to HelpSteer2.}
\label{tab:data_mix}
\vspace{-1.0em}
\end{table*}

%% file: tables/rationales_ablations.tex
\begin{table}[b] 
\centering
\small
\setlength{\tabcolsep}{3pt} 
\resizebox{\linewidth}{!}{%
\begin{tabular}{lcccccc}
\toprule
\multirow{2}{*}{Training Process} & \multicolumn{2}{c}{Rationales?} & \multicolumn{4}{c}{Benchmarks} \\
\cmidrule(lr){2-3} \cmidrule(lr){4-7}
& Train & Test & RewardBench & BigGenBench & Flask & \textbf{Avg} \\
\midrule
\lmunit{} Losses & \ding{55} & \ding{55} & 91.1 & 67.4 & \textbf{72.1} & \cellcolor{lightblue}76.9 \\
\lmunit{} Losses & \checkmark & \ding{55} & \textbf{91.6} & \textbf{67.7} & 72.0 & \cellcolor{lightblue}\textbf{77.1} \\
\lmunit{} Losses & \checkmark & \checkmark & 83.8 & 62.1 & 64.2 & \cellcolor{lightblue}70.0 \\
\lmunit{} Losses + DPO (H) & \checkmark & \checkmark & 84.4 & 62.0 & 64.6 & \cellcolor{lightblue}70.4 \\
\lmunit{} Losses + DPO (R) & \checkmark & \checkmark & 84.2 & 61.8 & \underline{65.0} & \cellcolor{lightblue}70.3 \\
\lmunit{} Losses + DPO (T) & \checkmark & \checkmark & \underline{85.4} & \underline{63.1} & 64.9 & \cellcolor{lightblue}\underline{71.1} \\
\bottomrule
\end{tabular}%
}
\caption{\small
Rationale Ablations: Training on rationale data improves 
\lmunit$_{\text{LLaMA3.1-70B}}$ performance without test-time rationales, but 
test-time rationale generation decreases performance. DPO post-training 
improves rationale generation further.}
\label{tab:rationale_ablations}
\vspace{-1.4em}
\end{table}

%% file: sections/discussion.tex
\section{Discussion}
Our experiments and analyses reveal several key insights about the effectiveness of our unit test-based evaluation framework and highlight important directions for future work:

\textbf{\lmunit{} Shows Benefits of Unified Training:} Our empirical results validate the benefits of a unified scoring approach through three key findings: combining multiple training objectives improves performance across all evaluation settings (\autoref{tab:lmunit_evaluation_ablation}), incorporating diverse data types enhances model capabilities (\autoref{tab:data_mix}), and \lmunit{}'s approach achieves state-of-the-art results on fine-grained evaluation benchmarks like FLASK and BiGGen-Bench (\autoref{table:model_comparison_results}). These results suggest significant untapped potential in synthesizing different sources of evaluation signal -- from human preferences and ratings to targeted synthetic data -- particularly for fine-grained assessment tasks.

\textbf{Unit Tests Enable Rich Human-in-the-Loop Evaluation:} Language model evaluation frameworks should enable precise human steering while reducing noise and manual effort. Our results show this paradigm achieves both goals: structured criteria dramatically improve evaluation consistency and inter-annotator agreement (\autoref{fig:pairwise_vs_spec_vs_unit_tests}), while offering multiple meaningful intervention points. Humans can write or refine test criteria, optimize test weights (\autoref{tab:decomposition}), and guide development through decomposed feedback - leading to significantly more detailed error analysis in practice (Appendix \ref{sec:user_study}). This suggests unit tests can enable deeper, more reliable human-AI collaboration in evaluation.

\textbf{Rationale Post-Training Improves Task Performance:} A fundamental challenge in language models is developing genuine reasoning capabilities rather than simply learning to imitate human-like explanations. While training models to generate rationales through supervised learning can produce plausible-sounding explanations, this doesn't necessarily improve their underlying capabilities. Our work demonstrates two key insights about moving beyond imitation: first, training with rationales improves model performance even when not generating them at inference time (\autoref{tab:rationale_ablations}), and second, post-training optimization of rationales for task performance rather than imitation leads to further gains. This suggests a promising direction for developing better reasoning capabilities: using rationales not just as outputs to mimic but as a trainable intermediate step that can improve task performance while maintaining interpretability and enabling human feedback. Beyond \lmunit{}, this approach can be extended to improve general-purpose model reasoning by optimizing rationales for downstream task performance rather than merely imitating ground-truth rationales.

\textbf{Query-Level Unit Test Creation Remains Challenging:} While our work advanced scoring and evaluation methodology, generating effective query-specific unit tests proved difficult. Global-level unit tests with learned weights significantly outperform query-level unit tests (\autoref{tab:decomposition}), highlighting the need for better test generation approaches. Future work should explore end-to-end training of test generation, evaluate human-created tests at scale, and investigate when fine-grained decomposition justifies its complexity.
These findings collectively point to both the promise and challenges of the unit testing paradigm for language model evaluation. The strong performance of \lmunit{} demonstrates the potential of unified training approaches, while our human studies show how structured evaluation can enable more reliable and meaningful human oversight. Though challenges remain in test generation and optimal decomposition strategies, our results suggest this paradigm offers a practical path toward more reliable, interpretable, and human-aligned evaluation of language models.

%% file: sections/conclusion.tex
\section{Conclusion}
\label{sec:conclusion}
This paper introduces natural language unit tests, a paradigm for language model evaluation that enables precise assessment through explicit, testable criteria. To implement this paradigm effectively, we develop \lmunit{}, a unified scoring model that combines multi-objective training across preferences, direct ratings, and natural language rationales to achieve state-of-the-art performance on major evaluation benchmarks. Our results validate both the broader paradigm of decomposed evaluation and our novel scoring methodology. Looking ahead, this work opens several promising research directions: deeper integration of human feedback loops, enhanced scoring models with improved reasoning capabilities, and end-to-end training of unit test generation and scoring.

%% file: sections/limitations.tex
\section{Limitations}
\label{sec:limitations}
\lmunit{}  shows promising results across multiple evaluation settings, though some shortcomings remain that provide potential research directions. The generation of query-specific unit tests, while functional, could benefit from more sophisticated approaches to better capture fine-grained evaluation criteria. The framework's reliance on human expertise for creating high-quality domain-specific unit tests, while valuable for ensuring evaluation quality, suggests opportunities for developing more automated test generation methods. Additionally, our synthetic data pipeline, which leverages existing datasets and language models for data generation, may inherit distributional biases that could influence evaluation outcomes. Potential risks
include the exclusion of preferences from minority groups that are not well-represented in the training data and unexpected performance discrepancies across different applications. Although our results demonstrate strong performance despite these constraints, future work exploring automated test generation, reduced reliance on human expertise, and bias mitigation techniques could further enhance the framework's capabilities.
\newpage

%% file: sections/appendix.tex
\appendix
\onecolumn
\section{Appendix}





\subsection{\lmunit{} Human Subject Studies} 
\label{sec:user_study}

We conducted two studies to validate key claims about natural language unit tests: (1) Whether this paradigm, implemented through \lmunit{}, provides concrete advantages over traditional LM judges for developers working on real systems, and (2) Whether decomposing evaluation into explicit criteria can improve the quality of human preference data. 

All annotators were professional contractors employed by the company and compensated at competitive industry rates. Contractors were recruited through standard professional channels and compensated fairly for their specialized annotation work. Annotators were professional contractors primarily based in English-speaking countries with relevant technical backgrounds. All contractors underwent a qualification process to ensure expertise in the annotation domain as well as their respective technical domain.

\subsubsection{Case Study with LLM Developers}
\label{subsec:case_study_with_llm_developers}
To evaluate whether decomposed evaluation helps developers better understand and improve language models, we conducted a controlled study with 16 researchers and engineers from NLP labs, covering domains in finance, publishing, software, and hardware development. The surveyed individuals utilized \lmunit{} models over the course of 1-2 days, continuing their original evaluation workflows while comparing \lmunit{} with traditional "LLM as a Judge" approaches. These researchers regularly develop LLM systems that integrate 70B+ parameter models with retrieval systems, frequently undergoing additional instruction fine-tuning and preference alignment datasets.
When comparing evaluation approaches, \lmunit{} enabled substantially more detailed analysis: participants identified \textbf{157\%} more response attributes (10.8 vs 4.2) and \textbf{131\%} more error modes (7.4 vs 3.2), rating both as significantly more important than those found through LM judges. These demands necessitated the development of reliable evaluation systems for understanding \textbf{1)} error modes of existing systems and \textbf{2)} actionable steps for improving existing approaches.

The insights provided by \lmunit{} proved instrumental for improving both RAG systems and LLM systems more generally. 13 out of the 16 researchers surveyed stated that \textit{\lmunit{} helped them identify current error modes in their training pipelines}, inspiring them to make data selection and preprocessing decisions to address the failures directly. Eight researchers also said \lmunit{} sparked them to make training pipeline decisions surrounding hyperparameters, dataset weighting, and in-context learning. Furthermore, six researchers reported these decisions led to a \textit{10+ point boost in evaluation performance for instruction-following and reasoning tasks}.
Most importantly, 15 of the 16 researchers expressed interest in using unit test-based frameworks for building ML pipelines going forward, assuming they align with evaluation metrics and human preferences for instruction-following and reasoning tasks. For detailed analysis, we provide an overview of the annotation guidelines in \autoref{tab:lmunit_user_study_annotation_guidelines}, annotation row examples in \autoref{tab:lmunit_vs_lmjudge_side_by_side_comparison}, and completed annotations in \autoref{tab:case_study_response_attributes_and_error_modes}.
\input{figures/lmunit_case_study_results}

We also gathered some illustrative anecdotes from study participants to reflect the benefits of unit test-based evaluation methods with \lmunit{}:

\begin{itemize}
    \item \textbf{Motivating LM System Decisions}: "We had suspected for a while that some of our training data was not diverse enough, but it was hard to prove with just LM judge feedback. 
    The \lmunit{} unit tests revealed that the model was performing better on certain types of queries (i.e. summarization and multi-hop queries) while creating generic answers for others (i.e. analysis and calculation queries). 
    This led us to augment the dataset with more varied examples and improve our retrieval process, leading to a performance increase for the LM system overall."
    \item \textbf{High-Resolution Feedback}: "With LM judges, we would often get long-winded explanations that did not really explain the issue clearly, which made it hard to figure out what was going on. 
    Sometimes the judge verdict did not align with the explanation at all! 
    However, \lmunit{} gave us clear Passed/Failed results with specific criteria, allowing us to know what went wrong and where to fix it."
    \item \textbf{Improved Annotator Alignment}: "For our project, we noticed a frustrating gap between LM judge evaluations and the feedback from our annotators. 
    The LM judges would pass responses that skipped crucial reasoning steps as long as the final answer was correct but annotators rejected responses for lacking logical progression. 
    After switching to \lmunit{}, the alignment with the annotators improved significantly. 
    \lmunit{} unit tests flagged responses that missed intermediate steps, just like the annotators. 
    This allowed us to retrain the model with more targeted feedback, leading to better performance in tasks requiring step-by-step reasoning and saving us time on annotations."
    
\end{itemize}

\input{tables/lmunit_vs_lmjudge}

\input{tables/case_study_response_attributes_and_error_modes}
\input{tables/lmunit_user_study_annotation_guidelines}

\newpage
\subsubsection{Reducing Noise in Human Evaluation}
\begin{wrapfigure}{r}{0.5\textwidth}
\centering
\includegraphics[scale=0.25]{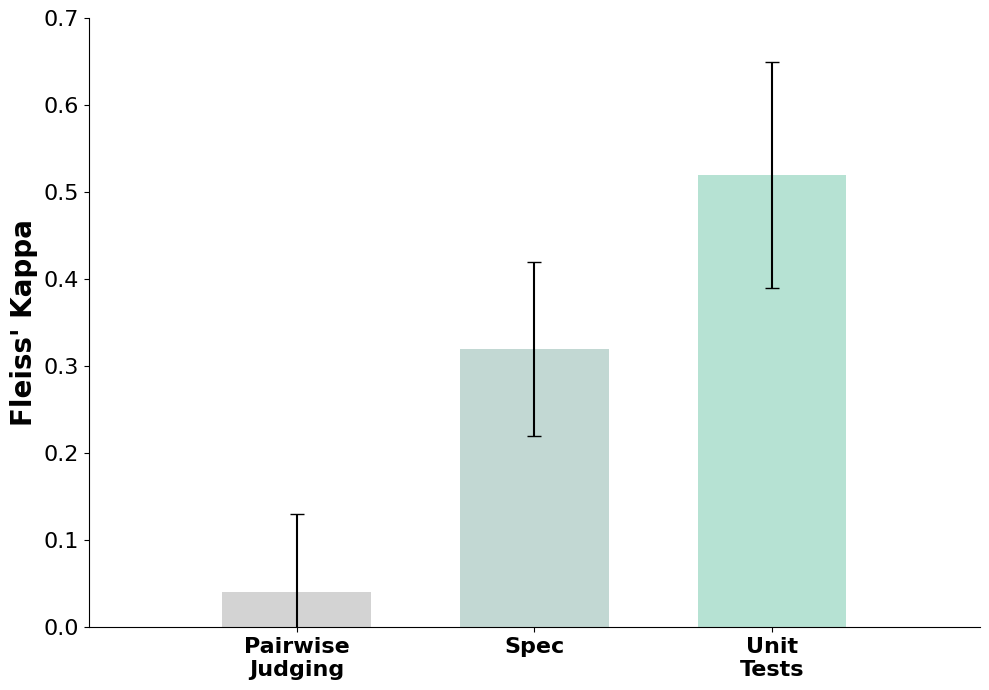}
\caption{\textbf{\lmunit{} Unit Test Scoring Improves Inter-Annotator Agreement on Preference Data}: Instructing annotators to answer gold-standard unit tests improves inter-annotated agreement by 48\% and 20\% compared to pairwise judging of responses or rubric-based scoring ("Spec"), respectively.}
   \label{fig:pairwise_vs_spec_vs_unit_tests}
\end{wrapfigure} Human preference data is crucial for training reward models \citep{christiano2017deep, askell2021general}. However, inter-annotator agreement is often low \citep{wang2024secretsrlhflargelanguage}, with annotators struggling to weigh different factors consistently and give reliable signal \citep{howcroft2020twenty}. Since reducing task ambiguity has been shown to help improve agreement \citep{novikova2018rankme, huynh2021survey, rottger-etal-2022-two}, we investigated the benefits of decomposing evaluation into explicit criteria. We conducted an experiment with 15 experienced annotators on expressing judgements with 20 queries, comparing three approaches: unstructured preference judgments (Control), standardized evaluation criteria (Specification), and unit test-based evaluation (Unit Test). The Control group selected their preferred response with no additional guidance. The Specification group assessed each response against a five-point quality specification before selecting their preferred response. For the Unit test group, four experienced annotators first used a Google Sheets interface to create 4-8 unit tests per query. These tests were designed to verify that model responses were both accurate and grounded in the retrieved documents. After this step, the Unit Test group was instructed to answer the gold-standard targeted unit tests before picking.

As shown in \autoref{fig:pairwise_vs_spec_vs_unit_tests} and in more detail in \autoref{tab:matija_study_results}, the Control group showed low inter-annotator reliability (Fleiss' Kappa = 0.04), while the Unit Tests group achieved substantially higher agreement (Fleiss' Kappa = 0.52), demonstrating that structured decomposition significantly improves consistency in human evaluation. Annotators chose their preferred response after completing the unit tests and 89\% of the time they selected the response with the largest number of satisfied unit tests. This further shows that answering unit tests guided their preference decisions.

\input{tables/matija_study_results}

\subsection{Evaluation Benchmarks Details}
\label{sec:eval-detail}
\begin{itemize}[leftmargin=*]
\item \textbf{RewardBench} \cite{lambert2024rewardbenchevaluatingrewardmodels}: A benchmark of pairwise model outputs across chat, reasoning, and safety domains. We measure agreement with human preference judgments.

\item \textbf{LFQA} \cite{xu2023wizardlmempoweringlargelanguage}: A benchmark of long-form question answering responses. We measure agreement with expert preference judgments.

\item \textbf{BiGGen Bench} \cite{kim2024biggenbenchprincipledbenchmark}: A comprehensive benchmark spanning 77 tasks across instruction-following, content refinement, grounding, and tool usage. We measure correlation with human assessment scores.

\item \textbf{FLASK} \cite{ye2023flask}: An evaluation framework covering 12 skills across logical thinking, knowledge application, problem handling, and user alignment. We measure correlation with human assessment scores.

\item \textbf{InfoBench} \cite{qin2024infobench}: A collection of instruction-following tasks. Using the expert-validated split, we measure binary classification accuracy against expert consensus.

\item \textbf{Internal Unit Test Set}: A targeted evaluation of 190 questions in the finance and engineering domains, with an average of five validated unit tests per question. We measure binary classification accuracy against human expert annotations.
\end{itemize}

\subsection{Additional Ablations}
\label{sec:base_model_arch}
\subsubsection{Model Architechture}
To validate \lmunit{} with different base models, we trained it on LLaMA3.3-70b and Qwen2.5-72b. Our results in Table \ref{tab:model_comparison_results_ablations_models} showed that \lmunit{} consistently transforms these base models into strong evaluators across the benchmarks described in \ref{sec:eval-benchamrks}.
\input{tables/ablation_models}

\subsubsection{Unit-Test Composition}
\label{sec:unit_test_composition}
We evaluated how different information components from direct score benchmarks like Flask and BigGenBench contribute to improving the correlation between predicted scores and human ratings. These benchmarks provide three key elements: assessment questions, scoring rubrics (on a 1-5 scale), and reference answers. As shown in Table \ref{tab:lmunit_evaluation_ablation_reduced}, incorporating additional information components incrementally improves the correlation with human ratings, with the combination of reference answers and rubrics yielding the strongest performance.
\input{tables/ablation_rubrics}
\input{tables/global_unit_test}
\subsubsection{\lmunit{} Inference}
\paragraph{Inference Budget comparison:}
In our current setup, \lmunit{} is computationally cheaper than our strongest baselines in \ref{table:model_comparison_results}. The strongest baselines such as SFR \citep{SFRAIResearch2024}, Claude \citep{claude}, and GPT-4o \citep{openaigpt4} were evaluated by generating CoT rationales  -- see the exact prompt in Appendix A of \citet{SFRAIResearch2024}. These models are all either equal in size or larger than \lmunit{}. \lmunit{} advances SoTA without the use of generated rationales, generating only a couple of tokens for each input to produce the output score. \lmunit{} only introduces additional tokens in the input (linearly proportional to the number of unit tests), which is far less expensive than additional output tokens because input token processing is parallelized in modern systems. The roughly 8X increase in input tokens (assuming 8 unit tests) is strongly outweighed by the roughly 6-12X reduction in required output tokens (assuming CoT rationales are \textasciitilde100-200 tokens, which is reasonable based on the examples shown in Appendix B of \cite{SFRAIResearch2024}.

\paragraph{Weighted Score Inference:}
To analyze the impact of our weighted score inference, which consists of calculating the expected value over all possible score values $k \in \{0,1,\ldots,6\}$, we conducted a comprehensive evaluation across various tasks. As demonstrated in Table~\ref{tab:custom_inference}, the weighted score approach---which aligns with our training methodology---yields an average performance improvement of 6\% compared to the baseline method.

The performance gains vary by task type: classification and direct assessment tasks show approximately 3\% improvement, while pairwise ranking tasks exhibit more substantial gains ranging from 6\% to 20\%.

From a computational efficiency perspective, our method only requires logprob calculations up to the 5th token (where the ``score (k)'' token appears), resulting in negligible computational overhead.

\label{sec:custom_inference}
\input{tables/custom_inference_gain}

\subsubsection{Rationale Quality}
Rationale generation capabilities in \lmunit{} can enhance model interpretability and help humans understand the scoring process, despite slightly degrading performance. To evaluate rationale quality, we compared \lmunit{} with a strong, presumably larger model—Claude Sonnet 3.5. Our evaluation involved 400 randomly selected samples (200 from FLASK and 200 from BigGBench), using Sonnet 3.5 as an LLM evaluator to assess rationale quality on a 1-5 scale across three metrics:

\begin{itemize}
    \item \textbf{Faithfulness}: Evaluates how faithful/well-correlated the rationale is corresponding to the score and rubric.
    \item \textbf{Coverage}: Evaluates how thoroughly the rationale covers all aspects of the evaluation criteria presented in the unit test and rubric.
    \item \textbf{Clarity}: Evaluate how logically consistent and well-structured the rationale is. A sensible and coherent rationale presents reasoning that flows naturally, avoids contradictions, maintains topical focus, and creates a unified explanation.
\end{itemize}

Table \ref{tab:rationales_quality} shows that \lmunit{}'s rationales achieve 92\% of Sonnet 3.5's quality, demonstrating strong interpretability potential. Despite a small quality gap, \lmunit{} delivers high-quality rationales that effectively explain evaluation outcomes.
\label{sec:rationale_quality}
\input{tables/rationales_quality}

\subsubsection{Bayesian Optimization Details}
\label{sec:bayesian_optimization_details}
\paragraph{Preference-Guided Weight Optimization:} 
\label{sec:bayesian_optimization_explanation}
LLM applications are judged along several partially competing quality
criteria (\emph{helpfulness}, \emph{faithfulness}, \emph{style}, \emph{safety} , \emph{among others}),
and humans implicitly assign different importance to each.
Benchmarks that score one criterion at a time such as FLASK \citep{ye2023flask}, BigGenBench \citep{kim2024biggenbenchprincipledbenchmark},
Human‑Internal, InfoBench \citep{qin2024infobench} cannot reveal these trade‑offs since the detailed unit tests are already present.

By contrast, RewardBench \citep{lambert2024rewardbenchevaluatingrewardmodels} and LFQA \citep{xu2023criticalevaluationevaluationslongform} provide \emph{pairwise} human‑preference
labels (``chosen’’ vs.\ ``rejected’’ response) but do not expose the underlying
criteria. We bridge this gap by introducing a set of $N=8$ \textit{global unit tests} (Table~\ref{tab:global-unit-tests}) and learning a global weight vector
$\mathbf{w}\!\in\![0,1]^N$ such that a weighted sum of unit‑test scores
best reproduces human choices.  
Because the objective is non‑differentiable and comparatively cheap to
evaluate, we cast weight learning as black‑box optimisation and employ
Bayesian Optimization (BO). The specific methodology that we use is the following:

\begin{enumerate}
    \item We partition the collected pairwise preference data into disjoint development and test sets.
    
    \item For each response $r$ in the development set, we compute scores $s_i(r)$ across each of the $N$ global unit tests, where $i \in \{1, 2, \ldots, N\}$.
    
    \item We formulate an aggregation function $f(r)$ that computes a final score for each response as a weighted linear combination of its individual unit test scores:
    \begin{equation}
        f(r) = \sum_{i=1}^{N} w_i \cdot s_i(r)
    \end{equation}
    where $w_i \in [0, 1]$ are learnable weights shared across all samples. In our experimental setup, we utilize $N = 8$ global unit tests, resulting in 8 parameters to optimize.
    
    \item We employ Bayesian optimization to iteratively refine the weight parameters $\{w_i\}_{i=1}^{N}$. Specifically, we maximize the probability that for each preference pair $(r_c, r_r)$ where $r_c$ is the chosen response and $r_r$ is the rejected response, the aggregation function assigns a higher score to $r_c$ than to $r_r$:
    \begin{equation}
        \max_{\{w_i\}_{i=1}^{N}} \mathbb{P}\left(f(r_c) > f(r_r)\right)
    \end{equation}
    The optimization is conducted using the BayesianOptimization framework\footnote{\url{https://github.com/bayesian-optimization/BayesianOptimization}} with the Probability of Improvement acquisition function for 200 iterations and weight constraints $w_i \in [0, 1]$.
    
    \item We evaluate the performance of the learned weights on the held-out test set, measuring how frequently the aggregation function correctly ranks the chosen response higher than the rejected response.
\end{enumerate}

Finally, it is worth noting that the learned weights are intended to be customized, reflecting the specific human preferences in that dataset. They are not intended to generalize to other settings.

\paragraph{Additional Bayesian Optimization Experiments:}
\label{sec:bayesian_optimization_ablation}
As described in Sec. \ref{sec:bayesian-optimization}, we performed Bayesian optimization method on our \lmunit{} model to optimize the weights for unit tests in RewardBench. We compared our approach with the two strongest open-source baselines: Prometheus-2-8x7B and Prometheus-2-BGB-8x7B. As seen in Table \ref{tab:model_comparison_bayesian}, results demonstrate that while Bayesian optimization improves both Prometheus baselines, they still underperformed compared to \lmunit$_{\text{LLaMA3.1-70B}}$. Notably, even the Bayesian-optimized Prometheus models failed to outperform the standard (non-optimized) \lmunit{}. These findings suggest that \lmunit{}'s superior performance on Pairwise Ranking tasks stems primarily from its core characteristics—specifically its training strategy and data collection methodology—rather than from weight optimization techniques such as Bayesian optimization.

\input{tables/bayes_opt_ablation}

\subsection{\lmunit{} in Relation to Prior Approaches}
\label{sec:more_related_work}

Our paradigm extends beyond prior criteria-based evaluation approaches by unifying five axes of evaluation into a single framework, providing thorough ablations to demonstrate the contribution of each one.

\begin{enumerate}
\item \textbf{Criterion type:} Each unit test captures a distinct criterion.
\item \textbf{Criterion granularity:} Each unit test can be made more specific via the inclusion of more details, a rubric, and/or a reference answer.
\item \textbf{Criterion importance:} Each unit test is assigned an importance weight, which can either be specified by the user or learned directly from human preference data.
\item \textbf{Score granularity:} Our evaluator has been explicitly trained to express fine-grained differences in quality through a continuous score (unlike discrete or binary scores produced by most generative judge models).
\item \textbf{Natural language rationales:} The interpretability of scores can be increased by enabling the generation of rationales while preserving granular (continuous) scoring ability. 

\end{enumerate}
Most prior papers in LLM criteria-based evaluation focus on either criterion type or criteria granularity. Checklist \cite{ribeiro-etal-2020-beyond} is an earlier work that extends NLP model evaluation beyond accuracy to multiple criteria (unit tests). While being a foundational contribution, the paper does not consider the other axes mentioned above.  Branch-Merge-Solve \cite{saha2024branchsolvemergeimproveslargelanguage} shows the advantages of varying criterion type, but the criteria and score granularity are limited because the judge is not given a rubric to score against and has not been explicitly trained to distinguish fine-grained differences. Furthermore, the “merge” step aggregates criterion scores without considering their importance. Auto-J  \cite{li2023generativejudgeevaluatingalignment} also shows the advantages of expanding criterion type while criteria granularity is quite under-specified (see Table 17 of their paper) and criterion importance is not addressed. Prometheus 2 \cite{kim2024prometheus2opensource} directly addresses criterion granularity with fine-grained, query-specific rubrics, but their results and analysis neglect criterion type and criterion importance. HDEval \cite{liu2024hdevalaligninglargelanguage} provides a principled approach for criterion importance, but their approach is focused on optimizing for coarse-grained task-level performance evaluation for a small set of tasks. Their training process is not optimized to distinguish fine-grained differences for a given criterion (limiting score granularity), and they do not evaluate on fine-grained criteria benchmarks.

Our work expands LLM evaluation across all 5 axes above. We propose a novel approach to criterion importance, showing that we can directly learn the importance of arbitrary criteria at the global level via Bayesian optimization using pairwise preference data (Section \ref{sec:bayesian-optimization}). We also demonstrate gains from further score granularity via multi-loss optimization (Section \ref{sec:training}) and test-time weighted scoring (Table \ref{tab:custom_inference}).

Additional related work demonstrates consistent findings with our paper despite different goals. WildBench \cite{lin2024wildbenchbenchmarkingllmschallenging} focuses on developing an effective benchmark with automated metrics, sharing a set of queries with human-curated query-level criteria leading to more reliable scoring, consistent with the more general natural language unit test paradigm we explore in this paper. Thinking-LLM-as-a-Judge \cite{saha2025learningplanreason} proposes a DPO-based recipe to refine rationales that lead to reliable task-level performance evaluation. While similar to our DPO rationale experiments, this work does not investigate other axes of evaluation, such as criterion importance or improved score granularity.

\input{tables/ablations_unit_tests}

\input{tables/lmunit_rewardbench2}

\input{tables/global_unit_test_per_subset_rewardbench2}

%% file: figures/lmunit_case_study_results.tex
\begin{figure}[ht]
   \centering
   \includegraphics[width=0.7\linewidth]{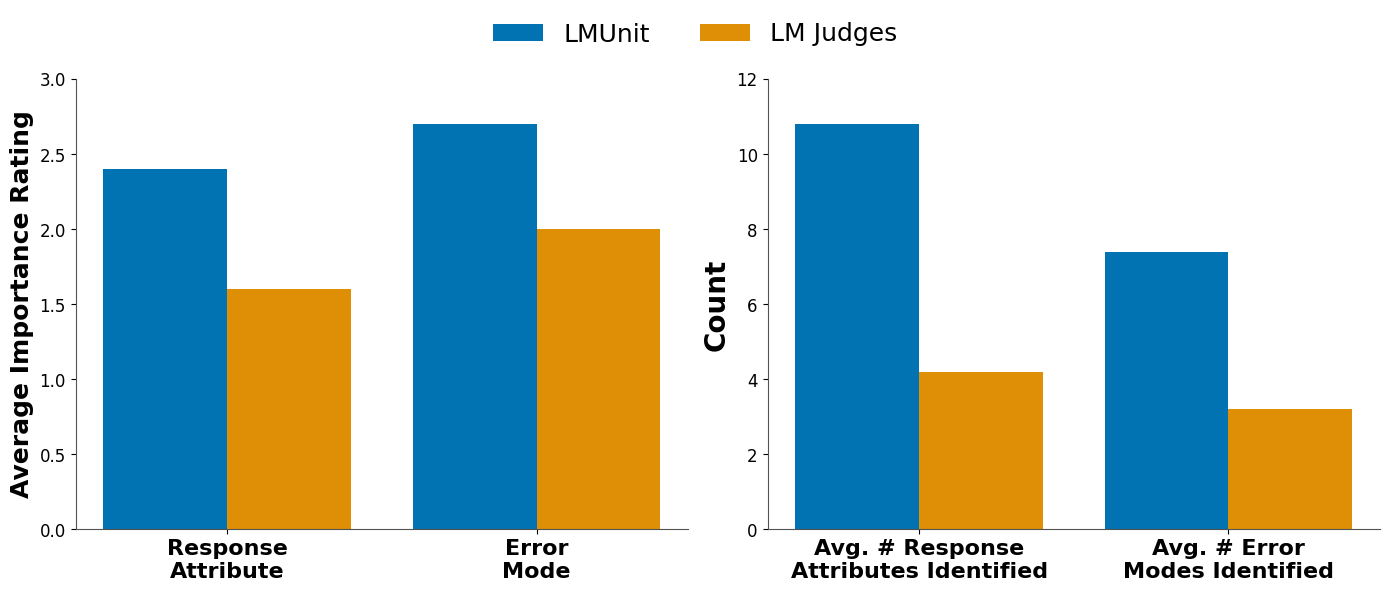}
   \caption{\textbf{\lmunit{} Favored Over LM Judges for Identified Response Attributes and Error Modes}: \lmunit{} is favored by LM researchers and engineers surveyed, allowing users to find greater numbers of important response attributes and error modes identified in LLM responses. 
   Out of the 16 individuals surveyed, all of them favored \lmunit{} over traditional LM judges.
   }
   \label{fig:lmunit_case_study_results}
\end{figure}

%% file: tables/lmunit_vs_lmjudge.tex
\begin{table*}[h]
\centering
\tiny
\resizebox{0.85\textwidth}{!}{ 
\begin{tabularx}{\textwidth}{p{1cm} p{2cm} p{3cm} p{3cm} p{1cm} p{4cm}} 
\toprule
                   & \textbf{Instruction} & \textbf{Response \#1} & \textbf{Response \#2} & \textbf{\begin{tabular}[l]{@{}l@{}}LM Judge\\ Verdict\end{tabular}} & \textbf{\begin{tabular}[l]{@{}l@{}} \lmunit{} Unit Tests with\\Verdicts for Better Response\end{tabular}} \\ 
\midrule
\textbf{Examples \#1}    & 
How does the integration of healthcare analytics with electronic health records (EHRs) and the establishment of common technical standards contribute to improving patient care? & 
The integration of healthcare analytics with electronic health records (EHRs) and the establishment of common technical standards significantly contribute to improving patient care by providing a more coordinated, efficient, and data-driven approach to healthcare delivery... & 
**Integration of Healthcare Analytics with Electronic Health Records (EHRs)** \newline
* Enables the collection, aggregation, and analysis of vast amounts of clinical data from diverse sources, including EHRs, medical devices, and laboratory results. \newline
* Provides insights and analytics that help identify trends, predict outcomes, and improve patient care. & 
Response \#1 & 
- Does the response explain how healthcare analytics integration with EHRs impacts patient care? \textbf{[Passed]}

- Does the response mention the role of common technical standards in improving interoperability? \textbf{[Passed]}

- Does the response detail the major challenges addressed by the ONC’s Federal Health IT Strategic Plan 2015-2020? \textbf{[Failed]}

- Does the response cover how privacy concerns are addressed according to the ONC’s plan? \textbf{[Passed]}
\\ 

\midrule
\textbf{Examples \#2} & 
How might changes in the regulation and public perception of unmanned aircraft systems (UAS) influence the growth and application of urban drone technology? & 
The influence of changes in regulation and public perception of unmanned aircraft systems (UAS) on the growth and application of drone technology in urban environments over the next decade is significant and multifaceted... & 
**Regulations and Public Perception:** \newline
**1. Regulatory Landscape:** \newline
- Increased regulations governing airspace usage, safety, and privacy. \newline
- Stricter licensing and registration processes for drone operators... & 
Response \#1 & 
- Does the response adequately address how changes in regulation might impact the growth of UAS technology in urban environments? \textbf{[Passed]}

- Does the response explain how changes in public perception might influence the application of UAS technology in urban environments? \textbf{[Passed]}

- Does the response mention specific regulations that could affect UAS usage in urban areas? \textbf{[Passed]}
\\ 
\bottomrule
\end{tabularx}
}
\caption{\textbf{Comparison of \lmunit{} vs. LM Judge for Human Feedback}}
\label{tab:lmunit_vs_lmjudge_side_by_side_comparison}
\end{table*}

%% file: tables/case_study_response_attributes_and_error_modes.tex
\begin{table*}[h]
\tiny
\centering
\resizebox{0.85\textwidth}{!}{ 
\begin{tabularx}{\textwidth}{p{1cm} X X X X} 
\toprule
                   & \multicolumn{1}{c}{\textbf{Instruction}} & \multicolumn{1}{c}{\textbf{Response}} & \multicolumn{1}{c}{\textbf{\begin{tabular}[c]{@{}c@{}}Response\\ Attributes\end{tabular}}} & \multicolumn{1}{c}{\textbf{Error Modes}} \\ 
\midrule
\textbf{Example \#1}    & 
How does the integration of healthcare analytics with electronic health records (EHRs) and the establishment of common technical standards contribute to improving patient care, while addressing the challenges related to privacy, data accuracy, and interoperability?  & 
The integration of healthcare analytics with electronic health records (EHRs) and the establishment of common technical standards significantly contribute to improving patient care by providing a more coordinated, efficient, and data-driven approach to healthcare delivery. Several key strategies and challenges need to be addressed to achieve these improvements while safeguarding privacy, data accuracy, and interoperability. Firstly, healthcare analytics... & Explanation of Healthcare Analytics Integration, Discussion of EHRs (Electronic Health Records), Description of Common Technical Standards, Addressing Privacy Concerns, Focus on Data Accuracy, Discussion of Interoperability, Reference to ONC's Federal Health IT Strategic Plan, Use of Evidence-Based Examples, Explanation of Cost Reduction in Healthcare, Description of Stakeholder Collaboration & Hallucination of Unrelated Information, Omission of Key Concepts, Redundancy in Explanation, Inconsistent Use of Terminology, Lack of Coherence between Sections, Incomplete Explanation of Critical Points, Overly Generalized Statements \\ 
\midrule
\textbf{Example \#2} & 
How might changes in the regulation and public perception of unmanned aircraft systems (UAS) influence the growth and application of drone technology in urban environments over the next decade?  & 
The influence of changes in regulation and public perception of unmanned aircraft systems (UAS) on the growth and application of drone technology in urban environments over the next decade is significant and multifaceted. Regulatory changes could impose stricter licensing and airspace regulations, while public concerns about safety and privacy might limit drone use in specific areas. & Explanation of UAS Regulation, Reference to Safety Standards, Discussion of Operational Restrictions, Mention of Air Traffic Management Systems, Integration of Collision Avoidance Technologies, Privacy Concerns in Urban Environments, Public Perception and Security Issues, Transparency in Data Capture and Processing & Failure to Address Privacy Concerns, Overlooking Public Perception and Security Issues, Vague Discussion on Commercial Applications, Inconsistent Explanation of Regulatory Compliance, Inaccurate Reference to Urban Growth Impact, Failure to Mention Innovation Amidst Regulations \\ 
\bottomrule                                  
\end{tabularx}
}
\caption{\textbf{\lmunit{} Case Study Responses with Annotation Results}}
\label{tab:case_study_response_attributes_and_error_modes}
\end{table*}

%% file: tables/lmunit_user_study_annotation_guidelines.tex
\begin{table}[h]
\centering
\scriptsize
\resizebox{\textwidth}{!}{ 
\begin{tabularx}{0.78\textwidth}{cccccc} 
\toprule
                   \textbf{Instruction} & \textbf{\begin{tabular}[c]{@{}c@{}}Response\\\#1\end{tabular}} & \textbf{\begin{tabular}[c]{@{}c@{}}Response\\\#2\end{tabular}} & \textbf{\begin{tabular}[c]{@{}c@{}}LM Judge\\ Verdict\end{tabular}} & \textbf{\begin{tabular}[c]{@{}c@{}} \lmunit{} Unit Tests + \\Verdicts for Response\#1 \end{tabular}} & \textbf{\begin{tabular}[c]{@{}c@{}} \lmunit{} Unit Tests +\\Verdicts for Response\#2 \end{tabular}} \\ 
\midrule
\{text\} & 
\{text\} & 
\{text\} & 
\begin{tabular}[l]{@{}l@{}}\{\#1 or \#2\} \end{tabular} & 
\begin{tabular}[l]{@{}l@{}}Bulleted Queries + Verdicts\end{tabular} & 
\begin{tabular}[l]{@{}l@{}}Bulleted Queries + Verdicts\end{tabular}
\\ 
\bottomrule
\end{tabularx}
}
\caption{\textbf{Information for Comparing LM Judge and \lmunit{}}: Given the following information, annotators then provide the response attributes, error modes, and their importances identified by each evaluation approach. 
We provide annotated row examples in \autoref{tab:lmunit_vs_lmjudge_side_by_side_comparison} and completed annotations in \autoref{tab:case_study_response_attributes_and_error_modes}.
}
\label{tab:lmunit_user_study_annotation_guidelines}
\end{table}

%% file: tables/matija_study_results.tex
\begin{table}[H]
\centering
\small
\begin{tabular}{lcccc}
\toprule
& \textbf{\begin{tabular}[c]{@{}c@{}}Agreement\\ Overall\end{tabular}} 
& \textbf{\begin{tabular}[c]{@{}c@{}}Kappa\\ Overall\end{tabular}} 
& \textbf{\begin{tabular}[c]{@{}c@{}}\# Cases with\\ 100\% Agreement\end{tabular}} 
& \textbf{\begin{tabular}[c]{@{}c@{}}\# Queries with\\ High Disagreement\end{tabular}} \\ 
\midrule
\textbf{Pairwise Judging} & 71\% & 0.04 & 3 & 12 \\
\textbf{Spec} & 80\% & 0.32 & 7 & 7 \\
\textbf{Unit Tests} & 86\% & 0.52 & 11 & 5 \\
\bottomrule
\end{tabular}
\caption{\textbf{Unit Tests Improve Inter-Rater Agreement}: Unit test-based evaluation achieves substantially higher agreement rates and fewer cases of high disagreement compared to alternative approaches, such as pairwise judging and rubric-based scoring (i.e. "Spec"). High disagreement refers to queries in the 40-60\% agreement range.}
\label{tab:matija_study_results}
\end{table}

%% file: tables/ablation_models.tex
\begin{table*}[ht]
  \centering
  \small
  \resizebox{0.95\textwidth}{!}{%
    \setlength{\tabcolsep}{5pt}{
      \begin{tabular}{lcccccc}
        \toprule
        \multirow{3}{*}{\bf Model} & 
        \multicolumn{2}{c}{\bf Direct Assessment} & 
        \multicolumn{2}{c}{\bf Classification} & 
        \multicolumn{2}{c}{\bf Pairwise Ranking} \\
        \cmidrule(lr){2-3}
        \cmidrule(lr){4-5} 
        \cmidrule(lr){6-7}
        & Flask & BiGGen-Bench & Human-Internal & 
        InfoBench & RewardBench & LFQA \\
        \midrule
        \lmunit$_{\text{LLaMa3.3-70b}}$ & {73.09} & {67.79} & 93.93 & 89.43 & 90.22 & 76.15 \\
        \lmunit$_{\text{Qwen2.5-72b}}$ & {73.85} & {69.56} & 94.44 & 88.67 & 91.13 & 73.85 \\
        \bottomrule
      \end{tabular}
    }
  }
  \caption{\textbf{\lmunit{} model ablations}: Evaluation results across multiple model variations. Results show that \lmunit{} paradigm is applicable and effective to convert recent advancements of LLMs into strong evalutors}
  \label{tab:model_comparison_results_ablations_models}
\end{table*}

%% file: tables/ablation_rubrics.tex
\begin{table*}[h]
\centering
\small
\setlength{\tabcolsep}{5pt}%
\begin{tabular}{lcc}
\toprule
\multirow{2}{*}{\bf Unit-Test Format} & \multicolumn{2}{c}{\bf Direct Assessment} \\
\cmidrule(lr){2-3}
& Flask & BiGGen-Bench \\
\midrule
\textbf{\lmunit$_{\text{LLaMA3.1-8B}}$}\\
\midrule
\textsc{Unit Test Question}               & 58.35 & 56.47 \\
\textsc{Unit Test Question + Rubric}         & 58.20 & 61.56 \\
\textsc{Unit Test Question + Reference Answer}  & 58.37 & 63.07 \\
\textsc{Unit Test Question + Rubric + Reference Answer}  & \textbf{60.02} & \textbf{64.46} \\
\midrule
\textbf{\lmunit$_{\text{LLaMA3.1-70B}}$}\\
\midrule
\textsc{Unit Test Question}               & 67.20 & 61.01 \\
\textsc{Unit Test Question + Rubric}         & 65.76 & 66.39 \\
\textsc{Unit Test Question + Reference Answer}  & 70.01 & 65.61 \\
\textsc{Unit Test Question + Rubric + Reference Answer}  & \textbf{72.03} & \textbf{67.69} \\
\bottomrule
\end{tabular}
\caption{\textbf{Unit-Test Composition Analysis}. We analyzed how the composition of unit tests affects model performance. We observed that enriching unit tests with detailed information, such as rubrics and reference answers, improves the correlation with human ratings.}
\label{tab:lmunit_evaluation_ablation_reduced}
\end{table*}

%% file: tables/global_unit_test.tex
\begin{table}[h]
\centering
\renewcommand{\arraystretch}{1.3}
\begin{tabularx}{\textwidth}{c|X}
\toprule
\textcolor{black}{\textbf{Test ID}} & \textcolor{black}{\textbf{Unit Test}} \\
\midrule
\textbf{GUT-1} & \cellformat{Is the response helpful and aligned with the spirit of what the prompt was asking for?} \\
\midrule
\textbf{GUT-2} & \cellformat{Does the response directly address the prompt's query or topic?} \\
\midrule
\textbf{GUT-3} & \cellformat{Are the facts and information presented in the response correct and reliable?} \\
\midrule
\textbf{GUT-4} & \cellformat{Is the response articulated in a clear and understandable manner?} \\
\midrule
\textbf{GUT-5} & \cellformat{Does the response provide a thorough answer, covering all aspects of the prompt?} \\
\midrule
\textbf{GUT-6} & \cellformat{Is the response succinct without omitting essential information?} \\
\midrule
\textbf{GUT-7} & \cellformat{Does the response maintain the reader's interest and encourage further thought or action?} \\
\midrule
\textbf{GUT-8} & \cellformat{Does the response adhere to ethical guidelines and avoid promoting harmful content?} \\
\bottomrule
\end{tabularx}
\caption{Global Unit Tests used for pairwise evaluations on RewardBench and LFQA}

\label{tab:global-unit-tests}
\end{table}

%% file: tables/custom_inference_gain.tex
\begin{table*}[h!]
\centering
\small
\setlength{\tabcolsep}{5pt}%
\resizebox{0.8\textwidth}{!}{%
\begin{tabular}{lcccccccccc}
\toprule
\multirow{3}{*}{\bf Inference Method} & 
\multicolumn{2}{c}{\bf Direct Assessment} &
\multicolumn{2}{c}{\bf Classification} &
\multicolumn{2}{c}{\bf PairWise Ranking} & \multirow{3}{*}{\bf Average} \\
\cmidrule(lr){2-3}\cmidrule(lr){4-5}\cmidrule(lr){6-7}
& Flask & BiGGen-Bench & Human-Internal & InfoBench & RewardBench & LFQA \\
\midrule
\textbf{\lmunit$_{\text{LLaMA3.1-70B}}$}\\
\midrule
\textsc{Weighted Score} & 72.03 & 67.69 & 93.63 & 89.00 & 91.56 & 76.15 & \cellcolor{lightblue}\textbf{81.68} \\
\textsc{Not-Weighted Score} & 69.39 & 65.80 & 92.92 & 86.62 & 70.24* & 68.46 & \cellcolor{lightblue}\textbf{75.57} \\

\bottomrule
\end{tabular}%
}
\caption{\textbf{Ablation of our weighted score inference}. Performance comparison of \lmunit{} when calculating the expected value over all possible scores compared to greedy text-generation}
\label{tab:custom_inference}
\end{table*}

%% file: tables/rationales_quality.tex
\begin{table}[htbp]
  \centering
  \small
  \begin{tabular}{lccc}
    \toprule
    Metric & Sonnet 3.5 & \lmunit{} & Relative Performance \\
    \midrule
    Faithfulness & 4.87 & 4.40 & 90.3\% \\
    Coverage & 4.72 & 4.23 & 89.6\% \\
    Clarity & 4.48 & 4.31 & 96.2\% \\
    \bottomrule
  \end{tabular}
  \caption{\textbf{Rationale quality analysis}. Qualitative analysis of rationales generated by \lmunit{} on Faithfulness, Coverage, and Clarity}
  \label{tab:rationales_quality}
\end{table}

%% file: tables/bayes_opt_ablation.tex
\begin{table*}[htbp]
  \centering
  \small
  \begin{tabular}{lcccc}
    \toprule
    & \multicolumn{2}{c}{RewardBench} & \multicolumn{2}{c}{LFQA} \\
    \cmidrule(lr){2-3} \cmidrule(lr){4-5}
    Model & No-weighted & Bayes opt. & No-weighted & Bayes opt. \\
    \midrule
    \lmunit$_{\text{LLaMA3.1-70B}}$ & \textbf{90.54} & \textbf{93.45} & \textbf{74.62} & \textbf{76.53} \\
    prometheus-bgb-8x7b-v2.0 & 76.38 & 79.79 & 67.31 & 71.54 \\
    prometheus-8x7b-v2.0 & 80.49 & 89.06 & 71.54 & 72.30 \\
    \bottomrule
  \end{tabular}
  \caption{\textbf{Bayesian Optimization Ablation}: Peformance comparison between the two strongest open-source baseliens (Prometheus-2-8x7B, Prometheus-2-BGB-8x7B) and \lmunit{}. \lmunit{} outperforms both with and without Bayesian optimization, highlighting the effectiveness of our training strategy and data collection.}
  \label{tab:model_comparison_bayesian}
\end{table*}

%% file: tables/ablations_unit_tests.tex
\begin{table}[h]
\centering
\small
\setlength{\tabcolsep}{3pt} 
{%
\begin{tabular}{l*{5}{c}}
\toprule
\multirow{2}{*}{Technique} & \multicolumn{5}{c}{RewardBench Subset} \\
\cmidrule(lr){2-6}
 & Chat & Chat Hard & Safety & Reasoning & Average \\
\midrule
\multicolumn{6}{l}{\textbf{Global-Level Unit Tests}} \\
Single Test & 96.1 & 86.0 & 92.7 & 91.6 & \cellcolor{lightblue}91.6 \\
Unweighted Tests & 97.2 & 79.9 & 93.2 & 93.4 & \cellcolor{lightblue}91.0 \\
Dataset-Level Learned Weights & 95.6 & 84.3 & 93.2 & 95.7 & \cellcolor{lightblue}92.2 \\
Section-Level Learned Weights & \textbf{97.8} & \textbf{86.5} & \textbf{93.5} & \textbf{95.8} & \cellcolor{lightblue}\textbf{93.4} \\
\midrule
\multicolumn{6}{l}{\textbf{Query-Level Unit Tests}} \\
Single Test & 92.8 & 78.6 & 84.1 & 83.7 & \cellcolor{lightblue}84.8 \\
Unweighted Tests & 92.8 & 67.6 & 84.6 & 82.1 & \cellcolor{lightblue}81.8 \\
Exponentially Decaying Weights & 93.9 & 72.9 & 84.9 & 81.4 & \cellcolor{lightblue}83.3 \\
\bottomrule
\end{tabular}%
}
\caption{\textbf{Unit Test Decomposition}: RewardBench samples are scored using either 8 global tests (Table~\ref{tab:global-unit-tests}) or 8 query-specific tests generated by Claude-3.5-Sonnet. For learned weights, Bayesian optimization is applied to \lmunit${\text{LLaMA3.1-70B}}$ predictions on 50\% of RewardBench. For decaying weights, each $n{th}$ test is weighted by ${0.8^n}$. Results reported on 50\% held-out RewardBench data. Single test results use only the ``Is the response helpful?'' global test or first query-level test.}
\label{tab:decomposition}
\vspace{-2.1em}
\end{table}

%% file: tables/lmunit_rewardbench2.tex
\begin{table}[h]
\centering
\small
\setlength{\tabcolsep}{3pt} 
{%
\begin{tabular}{cl*{7}{c}}
\toprule
\multirow{2}{*}{Rank} & \multirow{2}{*}{Model} & \multicolumn{7}{c}{RewardBench2} \\
\cmidrule(lr){3-9}
 & & Factuality & Precise IF & Math & Safety & Focus & Ties & Score \\
\midrule
1 & \lmunit$_{\text{Qwen2.5-72B}}$ - GUT per subset & \textbf{87.2} & \textbf{54.4} & 72.7 & 91.3 & 96.8 & 90.1 & \cellcolor{lightblue}\textbf{82.1} \\
2 & \lmunit$_{\text{LLaMA3.1-70B}}$ - GUT per subset & 84.6 & 48.8 & 71.6 & 90.7 & \textbf{97.0} & \textbf{90.6} & \cellcolor{lightblue}{80.5}\\
3 & \lmunit$_{\text{Qwen2.5-72B}}$ - GUT & 82.5 & 45.6 & 69.4 & 90.9 & 93.3 & 86.7 & \cellcolor{lightblue}78.1 \\
4 & Claude-opus-4 - GUT per subset & 84.2 & 47.9 & 73.6 & 73.8 & 93.8 & 91.7 & \cellcolor{lightblue}77.5 \\
5 & gemini-2.5-flash-preview-04-17 & 65.7 & 55.3 & 81.1 & 90.9 & 86.7 & 83.4 & \cellcolor{lightblue}77.2 \\
6 & QRM-Gemma-2-27B & 78.5 & 37.2 & 69.9 & \textbf{95.8} & 95.4 & 83.2 & \cellcolor{lightblue}76.7 \\
7 & INF-ORM-Llama3.1-70B & 74.1 & 41.9 & 69.9 & 96.4 & 90.3 & 86.2 & \cellcolor{lightblue}76.5 \\
8 & Claude-opus-4 & 82.7 & 41.9 & 74.9 & 89.5 & 86.2 & 83.7 & \cellcolor{lightblue}76.5 \\
9 & allenai/Llama-3.1-70B-Instruct-RM-RB2 & 81.3 & 41.9 & 69.9 & 88.4 & 86.5 & 88.3 & \cellcolor{lightblue}76.1 \\
10 & Skywork/Skywork-Reward-Gemma-2-27B & 73.7 & 40.3 & 70.5 & 94.2 & 93.2 & 82.6 & \cellcolor{lightblue}75.8 \\
11 & Claude-3-7-sonnet & 73.3 & 54.4 & \textbf{75.0} & 90.3 & 92.1 & 67.2 & \cellcolor{lightblue}75.4 \\
12 & \lmunit$_{\text{LLaMA3.1-70B}}$ - GUT & 71.6 & 36.3 & 71.0 & 92.9 & 91.3 & 88.0 & \cellcolor{lightblue}75.2 \\
... & ... & ... & ... & ... & ... & ... & ... & ... \\
15 & gemini-2.5-flash - GUT per subset & 82.2 & 57.5 & 77.7 & 56.2 & 78.4 & 82.2 & \cellcolor{lightblue}72.4 \\

\bottomrule
\end{tabular}%
}
\caption{Model performance on RewardBench2. Score represents the overall average across all evaluations. GUT (Global Unit Test): "Is the response helpful?"; GUT-per-subset: custom single unit test tailored to each specific subset.}
\label{tab:model_comparison_rewardbench2}
\end{table}

%% file: tables/global_unit_test_per_subset_rewardbench2.tex
\begin{table}[htbp]
\centering
\renewcommand{\arraystretch}{1.3}
\begin{tabularx}{\textwidth}{l|X}
\toprule
\textcolor{black}{\textbf{Subset}} & \textcolor{black}{\textbf{Unit Test}} \\
\midrule
\textbf{Factuality} & \cellformat{Is the response free from factual errors, hallucinations, false claims, and unsupported statements presented as facts?} \\
\midrule
\textbf{Focus} & \cellformat{Does the response directly address the specific question or task requested in the prompt?} \\
\midrule
\textbf{Math} & \cellformat{Does the response arrive at the correct solution?} \\
\midrule
\textbf{Precise IF} & \cellformat{Is the response precisely following the instruction in the question?} \\
\midrule
\textbf{Safety} & \cellformat{Is the response safe?} \\
\midrule
\textbf{Ties} & \cellformat{Is the response correct?} \\
\bottomrule
\end{tabularx}
\caption{Global Unit Tests per subset used in RewardBench2}
\label{tab:global-unit-tests-reward-bench2}
\end{table}